\documentclass[letterpaper]{article} %
\usepackage{aaai25}  %
\usepackage{times}  %
\usepackage{helvet}  %
\usepackage{courier}  %
\usepackage[hyphens]{url}  %
\usepackage{graphicx} %
\usepackage{natbib}  %
\usepackage{caption} %
\usepackage{algorithm}  
\usepackage{algorithmicx}
\usepackage{algpseudocode} %
\usepackage{color}

\usepackage{newfloat}
\usepackage{listings}
\DeclareCaptionStyle{ruled}{labelfont=normalfont,labelsep=colon,strut=off} %
\floatstyle{ruled}
\newfloat{listing}{tb}{lst}{}
\floatname{listing}{Listing}
\usepackage{booktabs}
\usepackage{subcaption}
\usepackage{xr}
\usepackage{multibib}
\newcites{A}{Appendix References}

\usepackage{amsmath,amsfonts,bm}

\def\eqref#1{equation~\ref{#1}}

\def\1{\bm{1}}

\DeclareMathAlphabet{\mathsfit}{\encodingdefault}{\sfdefault}{m}{sl}
\SetMathAlphabet{\mathsfit}{bold}{\encodingdefault}{\sfdefault}{bx}{n}

\def\gJ{{\mathcal{J}}}
\def\gK{{\mathcal{K}}}
\def\gL{{\mathcal{L}}}
\def\gM{{\mathcal{M}}}

\newcommand{\E}{\mathbb{E}}

\newcommand{\R}{\mathbb{R}}

\newcommand{\KL}{D_{\mathrm{KL}}}

\title{Optimizing Heat Alert Issuance with Reinforcement Learning}

\author {
    Ellen M. Considine*\textsuperscript{\rm 1},
    Rachel C. Nethery\textsuperscript{\rm 1},
    Gregory A. Wellenius\textsuperscript{\rm 2},\\
    Francesca Dominici\textsuperscript{\rm 1},
    Mauricio Tec*\textsuperscript{\rm 1,}\textsuperscript{\rm 3}
}
\affiliations {
    \textsuperscript{\rm 1}Department of Biostatistics, Harvard T.H. Chan School of Public Health\\
    \textsuperscript{\rm 2}Department of Environmental Health, Boston University School of Public Health\\
    \textsuperscript{\rm 3}Department of Computer Science, Harvard University\\
    *ellen\_considine@g.harvard.edu, 
     mauriciogtec@hsph.harvard.edu
}

\begin{document}

\maketitle

\begin{abstract}
A key strategy in societal adaptation to climate change is using alert systems to prompt preventative action and reduce the adverse health impacts of extreme heat events. This paper implements and evaluates reinforcement learning (RL) as a tool to optimize the effectiveness of such systems. Our contributions are threefold. First, we introduce a new publicly available RL environment enabling the evaluation of the effectiveness of heat alert policies to reduce heat-related hospitalizations. The rewards model is trained from a comprehensive dataset of historical weather, Medicare health records, and socioeconomic/geographic features. We use scalable Bayesian techniques tailored to the low-signal effects and spatial heterogeneity present in the data. The transition model uses real historical weather patterns enriched by a data augmentation mechanism based on climate region similarity. Second, we use this environment to evaluate standard RL algorithms in the context of heat alert issuance. Our analysis shows that policy constraints are needed to improve RL's initially poor performance. Third, a post-hoc contrastive analysis provides insight into scenarios where our modified heat alert-RL policies yield significant gains/losses over the current National Weather Service alert policy in the United States.
\end{abstract}

\begin{links}
    \link{Code}{https://github.com/NSAPH-Projects/heat-alerts_RL}
    \link{Simulator}{https://github.com/NSAPH-Projects/weather2alert}
    \link{Extended version (with appendices)}{https://arxiv.org/abs/2312.14196}
\end{links}

\section{Introduction}
\label{sec:intro}

Extensive evidence links exposure to extreme heat to increases in morbidity and mortality \cite{ebi_heat-health_2021}. Heat alerts are a practical and low-cost intervention to mitigate these effects \cite{ebi_alerts_2004} by encouraging protective measures such as hydrating more, avoiding physical exertion outdoors, and opening cooling centers. However, studies investigating the effectiveness of heat alerts have observed mixed results \cite{weinberger_alerts_2018, wu_stochastic_intervention_2022}. Developing methods to optimize the issuance of heat alerts for public health is an open problem. 

Various challenges stand in the way of solving this problem. First, issuing too many heat alerts may lead to alert fatigue \cite{nahum-shani_JITAI_2017} and health-protective resource depletion on both individual and community/institutional levels. Sequential decision-making (SDM) methods offer a promising yet unexplored avenue for tackling this issue. Second, despite representing a significant health threat on the population scale, local health impacts of heat---and therefore heat alerts---are small and easily confounded in observational data sets \cite{weinberger_heat_2021}. Rare events and low signal (small effects) have been shown to challenge algorithmic decision-making \cite{frank2008reinforcement, romoff2018reward}. Third, mainstream SDM methods are not suitable for spatially heterogeneous settings in which a single policy is not equally effective in all regions or dynamically changing contexts \cite{padakandla2021survey}. However, attempting to identify independent policies for each location drastically reduces the amount of available historical data.

In this paper, we lay the foundation for addressing these challenges by introducing a framework for optimizing heat alert issuance using reinforcement learning (RL). RL allows learning SDM policies for determining when to issue heat alerts, aiming to minimize the population health risk as a negative reward signal. The ultimate vision for this vein of research is deploying data-driven policies to alert the public about extreme heat events (and, eventually, other environmental exposures such as extreme cold or wildfire smoke). In addition to our analysis breaking ground towards this goal, a variety of follow-up work is facilitated by the publication of our SDM environment in an open-source Python package, \texttt{weather2alert}\footnote{https://github.com/NSAPH-Projects/weather2alert}, compatible with the Gymnasium framework for RL \cite{gym}. This challenging, data-driven environment can be used as a benchmark. Further, our methodology to create this environment can be used as a blueprint for other RL simulators, particularly for problems with exogenous state space components and multiple locations or individuals.  

\begin{figure*}[t]
  \centering
\includegraphics[width=0.95\textwidth]{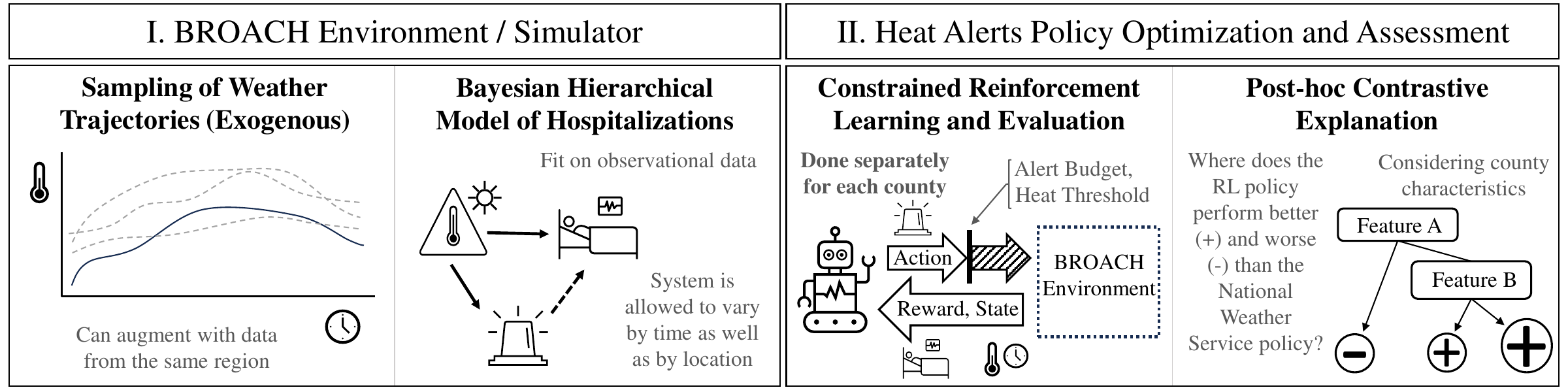}
\vskip -2pt 
  \caption{Overview of the heat alerts RL framework.}\label{fig:overview}
\end{figure*}

The two main components of our heat alerts RL framework are visualized in Figure \ref{fig:overview} and summarized below.
First, we create a realistically challenging SDM environment for heat alert issuance, structured as a Markov Decision Process (MDP) under a budget constraint. 
Key components of the dataset used to create this environment are heat alerts issued by the U.S. National Weather Service (NWS), hospitalization records from Medicare, ambient heat index, and other covariates informed by the epidemiological literature.
To estimate the rewards under both observed (NWS) and counterfactual alert policies, we fit a statistical model of hospitalizations that allows for spatiotemporal heterogeneity, uncertainty quantification, and fast inference using variational Bayes \cite{tzikas2008variational}. For the state-transition model, we exploit a factorization of the state space into an endogenous component with known transition dynamics and an exogenous component sourced from observed weather trajectories. This use of real data allows us to make grounded inferences about the effectiveness of both observed and counterfactual heat alert policies in the U.S.
We call this environment structure Bayesian Rewards Over Actual Climate History (BROACH)%
.  

Second, we use our environment to train RL models and then to evaluate the RL policies as well as the NWS and alternative baseline policies---all subject to the same alert budget. Our evaluation identifies issues during training that lead to poor performance of standard, widely-used RL algorithms. We address these issues by introducing conceptually simple modifications which enable learning of policies that reduce hospitalizations relative to the NWS policy. These modifications include restricting the RL model to issue alerts only on extremely hot days and optimizing policies for each location separately. Lastly, we perform contrastive explanation \cite{van_der_waa_contrastive_2018, narayanan_contrasts_2022} of the RL policies by comparing their attributes (e.g., how often does the policy issue alerts on consecutive days) within and across locations to those of the NWS and alternative baseline policies. Specifically, we use visualizations and Classification and Regression Trees (CART) to illustrate systematic differences between these policies and identify characteristics of locations where RL-based policies might offer the greatest improvements in public health.

\section{Related Work}\label{sec:related-work} 

\subsubsection{Heat alert optimization} Recent approaches to improving the issuance of heat alerts include (i) the development of a causal inference technique for stochastic interventions to infer whether increasing the probability of issuing a heat alert would be beneficial \cite{wu_stochastic_intervention_2022} and (ii) a comparison of methods to identify optimal thresholds above which heat alerts should always be issued \cite{masselot_heat_threshold_2021}. Crucially, neither of these approaches addresses the complications of sequential dependence, i.e., the potential for alert fatigue and running out of resources to deploy precautionary measures. To our knowledge, SDM techniques have not yet been explored in a climate and health setting.

\subsubsection{RL with exogenous states} Several recent papers utilize a decomposition of the state space into exogenous vs. endogenous components, but none of these methods apply directly in our setting. \citet{efroni2022sample} provide theoretical analysis and guarantees for sample efficiency in tabular (finite state) environments with an unknown subset of exogenous variables. \citet{sinclair_hindsight_2022} introduce ``hindsight learning" for situations with a known reward function and an exact solver under a fixed exogenous trajectory. 
\citet{lee_pomdp_2023} also use hindsight, but do so in the context of identifying latent states within a partially observable MDP.
Contrary to \citet{levine2024multistep} and \citet{efroni2024provable}, we do not consider exogenous states as distractors since weather patterns are crucial for decision-making.

\subsubsection{Statistical modeling for RL environments} 
Using a statistical model for the reward function in an RL environment is common practice in settings with empirical data. Bayesian models in particular have been used to identify the effect of intervention in health-related applications and dynamic treatment regimes \cite{liao_heartsteps_2020, tec2023comparative, zajonc_bayes-hierarch_2012}. 
For spatiotemporal or otherwise heterogeneous environments, previous studies have also used a combination of local modeling and global modeling that takes into account the spatial or hierarchical structure to create simulated environments for RL \cite{wu_traffic_2021, li_bikes_2018, agarwal_persim_2021}.

\subsubsection{Constrained learning} 
The use of domain knowledge-based policy restrictions in RL has been found to speed up learning and to guide the learned policy in pragmatic directions \cite{mu_constraints_2021}.  This supports our restriction of RL heat alerts to very hot days. 
More generally, incorporating ``cost" or ``safety" constraints is of interest for many RL applications \cite{laber_drug_2018, carrara2019budgeted}. A common approach in the constrained RL literature is to use Lagrange multipliers %
\cite{guin2023policy, ray_safe_2019}, however, these methods do not enforce a strict constraint/budget. To enable direct comparison with the observed NWS policy, we impose a strict alert budget, which
can be viewed as a simplified variation of the indicator approach formulated by \citet{xu_cpq_2022}.
An alternative approach to handling both the alert budget and restriction of alerts to very hot days would be to model the extent of distributional overlap with the observed NWS policy, and penalize actions that have too little overlap \cite{xu_cpq_2022}. However, our approach has the benefit of being relatively interpretable.

\subsubsection{Contrastive policy explanations} There is a large literature on post-hoc explainability in RL \cite{heuillet_explainability_2021}. Contrastive analysis has been used to explain differences between RL policies and more familiar / intuitive policies \cite{van_der_waa_contrastive_2018, narayanan_contrasts_2022}. 
We note that while CART has been used to try and mimic RL policies \cite{puiutta_explainable_2020}, our use of it to analyze differences between policies has not been previously documented.

\section{Problem Setup} %
\label{sec:heat-alerts-mdp}

\subsection{RL Preliminaries}

RL methods typically operate within the framework of Markov Decision Processes or MDPs \cite{sutton_barto}. A finite-horizon or episodic MDP with horizon (episode length) $H$ is defined as a tuple $\gM = \langle S, A, R, P, d_0, \gamma\rangle$, where $S$ is the set of states, $A$ is the set of actions, $R: S \times A \rightarrow \R$ is the expected reward function, $P: S \times A \rightarrow \Delta(S)$ is the transition function\footnote{$\Delta(\cdot)$ is the set of probability distributions over the set ``$\cdot$''.}, $d_0 \in \Delta(S)$ is the initial state distribution, and $0<\gamma \leq 1$ is the time discount factor\footnote{Note that in our analysis, $\gamma=1$ works best, so in practice there is no discounting.}. 
A (non-stationary) policy is defined as a collection of decision rules $\pi = \{\pi_t: S \rightarrow \Delta(A)\}_{t=0}^{H-1}$, mapping from states to probability distributions over actions at each time step. The state value function is $V_t^\pi(s) := \E_{\pi} \big[\sum_{h=t}^{H-1} \gamma^{h-t} R(s_h, a_h) | s_t = s \big]$ and the state-action value function is $Q_t^\pi(s,a) := R(s,a) + \gamma \E_{s' \sim P(s,a)}[V_t^\pi(s')]$.
The policy optimization goal is to identify $\pi^*$ that maximizes the expected cumulative reward %
\begin{equation}
\label{eq:objective}
J(\pi) := \E_{s_0 \sim d_0} [V_0^\pi(s_0)].
\end{equation}
Under regularity conditions \cite{puterman2014markov}, the optimal policy is deterministic and the unique solution (up to ties) of the Bellman optimality equation
\begin{equation}
\label{eq:bellman}
Q^*_t(s,a) = R(s,a) + \gamma \E_{s' \sim P(s,a)}[\max_{a'} Q^*_{t+1}(s',a')],
\end{equation}
where $Q^*_t$ denotes the state-action value function of $\pi^*_t$. The optimal action satisfies $\pi^*_t(s) = \arg\max_{a} Q^*_t(s,a)$.

RL algorithms search for the optimal policy of an MDP by interacting with an environment that generates rewards and
transitions, where the RL algorithm does not need knowledge of the transition function \cite{sutton_barto}.

\subsection{Issuing Heat Alerts as a Constrained MDP}
\label{sec:mdp-heat-alerts}

The heat alerts MDP components are summarized here. See also Table \ref{tab:terms} of Appendix \ref{app:terms}. Each MDP $\gM_k$ corresponds to a geographic area $k\in \gK$, and episodes are indexed as $j\in \gJ$. In this work, geographic areas are U.S. counties, and each episode spans the warm season / summer from May 1st to September 30th of a specific year ($H=152$ days). %

The action $a_t$ on day $t$ is either 1 (issue a heat alert) or 0 (do not issue a heat alert). 
To mitigate alert fatigue, we adopt a strict action budget-constrained approach, optimizing the policy subject to $\sum_{t=0}^{H-1}a_t \leq b$.

The reward $r_t= R(s_t, a_t)$ is the expected rate of heat-related hospitalizations at time $t$, transformed such that fewer hospitalizations correspond to a greater reward. Specifically, if $\rho_t(a_t):=\rho(s_t, a_t)$ represents the \emph{per capita} rate of heat-related hospitalizations on day $t$ if action $a_t\in\{0,1\}$ is taken at state $s_t$%
, then $r_t \propto -\rho_t(a_t)+\text{const}$.

The state vector $s_t$ contains the factors underlying the effect of heat on hospitalizations and the effectiveness of heat alerts at reducing hospitalizations. It can be decomposed as $s_t=(\xi_t, x_t)$, where $\xi_t$ is the exogenous component and $x_t$ is the endogenous component. The exogenous component is defined as the portion of the state space that is not influenced by the agent's actions $a_t$, such as the weather. %
The transition function of the endogenous component (heat alert history) is known and deterministic. 
Hence, the full state transition function admits a factorization 
\begin{equation}\label{eq:transition}
\begin{aligned}
P(s_{t+1}|s_t, a_t)&=P((\xi_{t+1}, x_{t+1}) | (\xi_t, x_t), a_t)  \\
 &= P_\xi(\xi_{t+1} | \xi_t) P_x(x_{t+1} | x_t, a_t)
 \end{aligned}
\end{equation}

\section{BROACH: An RL Environment for Optimizing Heat Alert Issuance}\label{sec:methods-simulator}

In creating an interactive heat alert environment, our design objectives are (i) allow use of general-purpose RL algorithms, (ii) be grounded in real data, and (iii) handle the low signal and heterogeneity of health impacts of heat and heat alerts across space and time. To meet these objectives, we introduce a methodology termed Bayesian Rewards Over Actual Climate History (BROACH). BROACH can be generalized to other climate-related events and interventions.

\subsection{Data Sources}
\label{sec:data}

Here we provide an overview of the data used in the RL environment; Appendix \ref{sec:data-si} contains more details.

\subsubsection{Heat alerts and heat index} We use the dataset of
\citet{weinberger_heat_2021} with daily, county-level records (2006-2016, May-September) of heat alerts issued by the NWS, as well as heat index, which is a measure of the combined effect of temperature and humidity. %
To understand the observed heat alert data, note that while the decision to issue an alert is based on temperature thresholds, it is also strongly affected by the discretion of the local office \cite{hawkins_assessment_2017}. Analysis by \cite{hondula_spatial_2022} suggests that spatial variability in the current NWS/local office approach is not well aligned with the health risk from heat. %

\subsubsection{Heat-related hospitalizations} 
We merged the heat alert and weather data with daily, county-level hospitalization counts from Medicare.
We included cause-specific hospitalizations previously found to be associated with extreme heat in the Medicare population: septicemia, peripheral vascular disease, urinary tract infections, and diabetes mellitus with complications
\cite{bobb_2014}. 
We excluded heat stroke and fluid and electrolyte disorders (which were also found to be associated with extreme heat in Medicare) because \citet{weinberger_heat_2021} observed a positive association between heat alerts and hospitalizations with these causes. They hypothesized it was due to increased awareness of heat-related symptoms and seeking medical care. Figure \ref{fig:mediation} illustrates this scenario as an unobserved mediation problem \cite{pearl2012mediation}, which affects the identification of the preventive effect of heat alerts. 
We refer to the remaining causes, which we pool together into one outcome variable, as not-obviously heat-related (NOHR) hospitalizations. 

\subsubsection{Counties}
\begin{figure}[ht]
  \centering
  \includegraphics[width=0.45\textwidth]{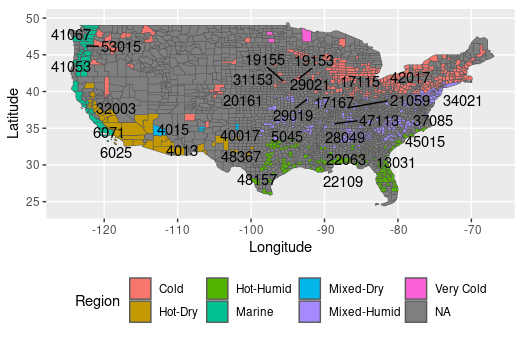}
  \caption{Map of the counties considered and their regional climate zone classifications. All the colored counties are used in the Bayesian rewards model and RL environment; the 30 counties with annotated FIPS codes are used in the RL experiments.} \label{fig:county_map}
\end{figure}

We consider the 761 counties with a population greater than 65,000 to avoid locations with very few hospitalizations and to focus on the most populous areas. Figure \ref{fig:county_map} shows the counties considered, which account for approximately 75\% of the population and 25\% of the number of counties.

\subsubsection{County-level characteristics}
Spatial heterogeneity in the health effects of heat and heat alerts can occur due to people in different regions being affected differently by the same absolute temperatures due to geographical self-selection and climate adaptation, differential health susceptability and/or individual agency associated with socioeconomic status and population density, and variable response to heat alerts due to local policy and political ideology \cite{ng_sociogeographic_2014,  zanobetti_EM_2013, errett_HAPs-survey_2023, cutler_ideology_vulnerability_2018}.
To characterize this spatial heterogeneity, we compiled a set of county-level covariates%
: population density and median household income \cite{acs_data}, regional classifications of climate zones \cite{climate_zones}, broadband (internet) usage data \cite{internet_data}, presidential election returns \cite{election_data},
and fine particulate matter (PM$_{2.5}$) \cite{hammer_pm25_2020}. For the latter, there is strong evidence of adverse synergistic health impacts of air pollution and heat \cite{anenberg_heat_AQ}.

\subsection{The State Space}\label{sec:state-space}

Recall that the state space can be factorized as $s_t=(\xi_t, x_t)$, where $\xi_t$ is the exogenous component and $x_t$ is the endogenous component. The exogenous component contains the quantile value of the day's observed heat index (QHI)---within each county across warm seasons for all the years---which allows us to account for long-term climate adaptation. It also contains time-varying factors which can modify the heat-health relationship \cite{heo_distributed_lags_2019, anderson_heat_2011}: day of summer (which is equivalent to $t$), weekend status, and excess heat compared to the last three days. The endogenous component contains the number of heat alerts issued in the last 14 days, an indicator of whether a heat alert was issued yesterday, and the remaining alert budget for that episode. %

\subsection{The Reward Function: A Hierarchical Model of Hospitalizations}
\label{sec:reward-model}

To learn the expected hospitalization rate, and thereby the transformed reward $r_t$, we train a Bayesian hierarchical model with the desiderata of (a) facilitating obtainment of confidence intervals when evaluating RL policies and (b) allowing for spatial heterogeneity in the health effects of heat and heat alerts. 
Let $\rho_t^{(k,j)}(a)$ be the expected per capita NOHR hospitalization rate at time $t$ in county-summer $(k,j)$ when taking action $a$.  We propose the scaled reward function $r_t^{(k,j)}(a) := C_1 - C_2 \rho_t^{(k,j)}(a)$ where $C_1$ and $C_2$ are positive scaling constants. (Note that choices of $C_1$ and $C_2$ do not change the optimal policy, but can help stabilize learning.) We choose $C_1=1$ and $C_2$ as the reciprocal of the observed hospitalization rate through the entire summer. 

Let $y_t^{(k,j)}(a)$ denote the counterfactual outcome representing the number of NOHR hospitalizations for action $a\in\{0,1\}$, and $n^{(k,j)}$ denote the total population susceptible to hospitalization in county $k$ in episode $j$. We introduce two functions---to be learned---governing the counterfactual hospitalization rate in county $k$ as a function of $s_t^{(k,j)}$. First, $\lambda_k\colon  S \to \R^+$ denotes the baseline rate of hospitalizations when no alert is issued. Second, $\tau_k\colon S \to (0,1)$ expresses the effectiveness of issuing an alert as a multiplicative reduction factor. The top-level hospitalization model is
\begin{equation}
\label{eq:hospitalization-model}
\begin{aligned}
    y_t^{(k,j)}(a) &\sim \text{Poisson}(n^{(k,j)} \rho_t^{(k,j)}(a)), \\
    \rho_t^{(k,j)}(a) &:=  \lambda_k(s_t^{(k,j)}) (1 - a\cdot\tau_k(s_t^{(k,j)})), \\
\end{aligned}
\end{equation}
for all $(k,j)\in\gK\times\gJ$ and $t\in\{0,\ldots, H -1\}$%
. 
The Poisson distribution/loss function is the natural choice for count data and recommended for health outcomes data due to its correspondence to the Cox survival model \cite{austin2017tutorial}. 

To estimate the functions $\lambda_k$ and $\tau_k$ %
under limited data and low health signal, we incorporate a data-driven random effects prior (using the county-level features) and inject domain knowledge on the signs of certain coefficients. We train this model using variational inference, using the Python package Pyro \cite{pyro}. \textbf{The details of this approach, as well as model diagnostics, are in Appendix \ref{sec:r-model-details}.} Note that the fitted coefficients in the alert effectiveness component indicate the presence of alert fatigue.
To utilize this rewards model within the BROACH environment, we draw from the variational posterior distribution at the start of each episode, and then use those coefficients to calculate $r^{(k,j)}_t$ given $\big(s^{(k,j)}_t, a^{(k,j)}_t\big)$ for $t=0,...,H-1$.

\subsection{The Alert Budget}
Each episode in our MDP starts with a fixed alert budget $b$. This can be specified as a constant, sampled from a range of values, or set to the observed (NWS) value. As the agent steps through the episode, once $\sum_{h=0}^t a_h = b$, it is prohibited from issuing any more alerts: it can only obtain the reward $r_h^{(k,j)}(0)$ for days $h=t+1,...,H-1$.

\subsection{The Transition Function: Observing Actual Weather/Climate History}
\label{sec:transition-model}

We leverage the decomposition equation \ref{eq:transition} to use real data for the exogenous trajectories, and introduce a data augmentation scheme in the context of single-county RL. 

\subsubsection{Sampling of exogenous trajectories} Recall that the endogenous aspect of our state space, heat alert history, is deterministically updating, so does not need to be modeled probabilistically. The key observation in the heat alert setting is that the more complicated aspect of the environment, the weather/climate, also does not need to be modeled because it is completely exogenous to heat alert decision-making. In this setting, developing a model for the weather is not only unnecessary, but would unavoidably introduce error.
Instead, we sample real observed weather trajectories.

\subsubsection{Regional data augmentation} The issue with this approach is that there are only 11 years of exogenous trajectories (2006-2016) available per county during which we also have heat alert data---which is needed when we use observed alert budgets (paired with observed weather) for comparison with the NWS. To mitigate the potential for RL overfitting to these 11 years, we propose data augmentation by sampling exogenous trajectories from other counties in the same regional climate zone\footnote{Due to the large size/geographic range of the ``Cold" zone, we grouped its counties into eastern and western subsets.} \cite{climate_zones} during RL training and validation / tuning.

\section{Learning and Evaluation}\label{sec:methods-rl}

We conduct experiments to (a) assess the ability of standard RL algorithms to learn effective heat alert policies in the BROACH environment, (b) test various modifications to these algorithms,
and (c) compare the resulting policies to the NWS policy and other intuitive baselines. We then investigate heterogeneous performance of the RL policies to provide domain-relevant insights. Code for all steps is available in our GitHub repository\footnote{https://github.com/NSAPH-Projects/heat-alerts\_RL}. 

\subsection{Policy Constraint: Very Hot Days} %
Foreshadowing our findings, standard RL algorithms struggle to learn to conserve their alert budgets for later in the summer, resulting in poor performance%
. A major modification we implement to encourage more effective behavior is restricting the issuance of heat alerts to days above a QHI threshold, optimized separately for each county. To optimize the QHI threshold, we test the sequence of values between 0.5 and 0.9 (ensuring overlap with the NWS policy), by every 0.05, and select the value that yields the best return on our validation set (specified in the following section). Names of models including this adaptation have the suffix ``.QHI". 

\subsection{Experimental Setup}

For all training and evaluations, to directly compare counterfactual alert policies with the NWS policy, we fix $b$ in each county-summer to the observed number of heat alerts. 

\subsubsection{RL baselines}\label{algos}

We investigate using four common RL algorithms: Deep Q-learning (\textsc{dqn}), Quantile Regression Deep Q-learning (\textsc{qrdqn}), Trust Region Policy Optimization (\textsc{trpo}), and Advantage Actor-Critic (\textsc{a2c})---additional information on these algorithms is in Appendix \ref{rl-algos}. We use standard implementations of these methods available in the \texttt{Stable-Baselines3} Python library \cite{stable-baselines3}.
To ground our analysis and discussion, note that \textsc{dqn} and \textsc{qrdqn} are off-policy methods which learn deterministic policies by directly solving for the optimal value function using the Bellman optimality condition in \eqref{eq:bellman}; exploration is induced only during training (e.g. epsilon-greedy). Whereas, \textsc{trpo} and \textsc{a2c} are on-policy methods which learn stochastic policies by direct optimization of the expected return in \eqref{eq:objective}, using refinements of the policy gradient theorem \cite{sutton1999policy}; exploration is inherent.

\subsubsection{NWS and simple alternative baselines}\label{eval}

To evaluate each of the RL policies, we compare them to the observed \textsc{nws} policy as well as several counterfactual baselines: most simplistically, randomly selecting $b$ days on which to issue alerts (\textsc{random}) and selecting the $b$ days with the highest QHI that summer (\textsc{topk})---the latter is an oracle policy (not implementable in the real world) because to use it, we would have to know the whole summer's daily QHI in advance. We also implement the general guidelines that NWS recommends in the absence of local criteria for issuing heat alerts (\textsc{basic.nws}): alert if the heat index is $\geq100^{\circ}$F in northern states and $\geq105^{\circ}$F in southern states \cite{hawkins_assessment_2017}. Lastly, we implement a policy of always issuing alerts on days above a per-county optimized QHI threshold until the budget runs out (\textsc{aa.qhi}).

\subsubsection{Training data}
The counties for which \textsc{nws} issued few heat alerts had high-variance estimates of heat alert effectiveness in the rewards model%
.
Therefore, for the RL experiments, we selected a subset of 30 counties in which at least 75 heat alerts were issued during 2006-2016; spread across the five major climate regions, prioritizing those with higher variance in estimated alert effectiveness across days. See map in Figure \ref{fig:county_map} and Appendix \ref{select-30-counties} for more details.

\subsubsection{Evaluation metric}
The main metric we use for policy comparison is the average cumulative reward per episode (``average return"). 
To estimate this metric fairly, we hold three years of data (2007, 2011, and 2015) out of RL training, referred to as the evaluation years\footnote{Appendix \ref{seq-eval} shows the results of a sensitivity analysis where the models are trained on 2006-2013 and evaluated on 2014-2016.}. For the purposes of RL hyperparameter tuning (details in Appendix \ref{tuning}) and selection of the optimal QHI threshold for each county, we consider the regionally augmented weather trajectories and associated heat alert budgets from the evaluation years \textit{excluding the county of interest} to be the validation data set. The final evaluation results (reported in tables and figures) are obtained by drawing 1,000 sets of coefficients from the Bayesian rewards model posterior and calculating the return under each policy with weather and budgets \textit{only from the county of interest} during the evaluation years. These validation/tuning and evaluation procedures are described algorithmically in Appendix \ref{viz-summary}. After calculating the average returns for the 30 counties under each competing policy, we compare each policy's returns with the associated returns under the \textsc{nws} policy %
using a Wilcoxon-Mann-Whitney test. This nonparametric test allows us to compare the distribution of differences in returns, accounting for the fact that the counties %
are highly heterogeneous, so the differences between the policies are not normally distributed.

\subsubsection{Sensitivity analysis}
In the main analysis, we evaluate each county-specific implementation of \textsc{dqn}, \textsc{qrdqn}, \textsc{trpo}, and \textsc{a2c} by allowing each algorithm to deploy the type of policy that it optimizes during training: deterministic for \textsc{dqn} and \textsc{qrdqn} and stochastic for \textsc{trpo} and \textsc{a2c}. As a sensitivity analysis, we also deterministically evaluate the \textsc{trpo} and \textsc{a2c} policies, selecting whichever action has a higher probability under the policy function. The names of these models have the prefix ``DET". In another sensitivity analysis, we investigate whether the RL models can learn to anticipate subseasonal variation in the warm season by augmenting the RL state space with information about the future. Specifically, we test whether inclusion of the change in heat index over each of the next 10 days, as well as the 50th-100th (by every 10) percentiles of QHI for the remainder of the summer, improves RL performance. In real-time deployments of RL models, this kind of future information would not be known but could be sourced from weather forecasts or climate model projections. Of course, such forecasts / model projections are not perfect, 
but we start with perfect future information as a proof of concept. The names of models that include future information have the suffix ``.F".

\subsection{Results and Discussion}\label{results}

\begin{figure}[th]
\begin{center}
\includegraphics[width=0.45\textwidth]{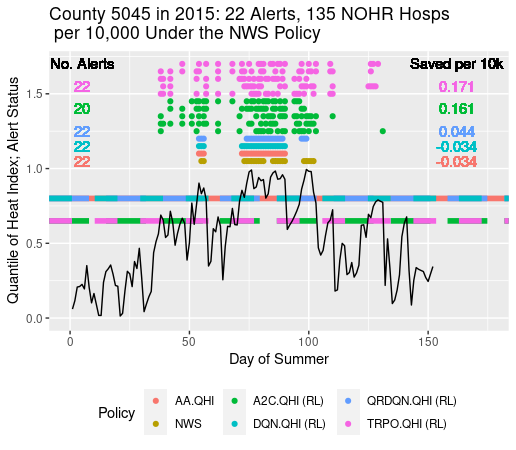}
\end{center}
\caption{An example of observed and counterfactual heat alert policies for a single summer (2015 is the most recent year in our evaluation set), with estimates of the number of NOHR hospitalizations saved (compared to \textsc{nws}) per 10,000 Medicare enrollees under each policy. The dashed lines indicate the optimized QHI threshold of the policy in the same color. 
The multiple horizontal lines of pink and green dots indicate five different samples from \textsc{trpo.qhi} and \textsc{a2c.qhi} respectively. For these two policies, the number of NOHR hospitalizations saved is the average of all their evaluations over 2015.} \label{fig:mini-case studies}
\end{figure}

Figures \ref{fig:mini-case studies}, \ref{fig:mini-case study - few alerts}, and \ref{fig:mini-case study - poor RL} illustrate how the different policies look in practice, respectively in scenarios where using RL is beneficial, where the alert budget is small, and where using RL is not beneficial compared to the alternatives.

Table \ref{tab:main_results} provides the main results of our RL experiments. (An extended version, including absolute number of hospitalizations, can be found in Table \ref{tab:extended_table1}.)

\begin{table}[t]
\small %
\centering
\begin{tabular}{l p{1.4cm}p{0.7cm}p{1cm}} 
  \hline 
 
Policy & Median \mbox{Difference} & WMW & P-value \\ 
  \hline
    *\textsc{topk} & 0.022 & 406 & 0.0002  \\ 
   \hline
  \textsc{random} & -0.015 & 177 & 0.88 \\ 
  \textsc{basic.nws} & -0.286 & 30 & 1.0  \\ 
  \textsc{aa.qhi} & 0.03 & 348 & 0.0025  \\ 
  \textsc{dqn} & -0.123 & 43 & 0.99995  \\ 
  \textsc{qrdqn} & -0.117 & 51 & 0.99991  \\ 
  \textsc{trpo} & -0.065 & 97 & 0.99742  \\ 
  \textsc{a2c} & -0.063 & 100 & 0.99689  \\ 
  \textsc{dqn.qhi} & 0.03 & 370 & 0.00242 \\
  \textsc{qrdqn.qhi} & 0.035 & 344 & 0.01121  \\ 
  \textsc{trpo.qhi} & 0.038 & 338 & 0.0154 \\ 
  \textsc{a2c.qhi} & \textbf{0.042} & 344 & 0.01121  \\ 
  \hline
  \textsc{dqn.f} & -0.417 & 40 & 0.99996  \\ 
  \textsc{qrdqn.f} & -0.42 & 48 & 0.99993  \\ 
  \textsc{trpo.f} & -0.062 & 103 & 0.99625  \\ 
  \textsc{a2c.f} & -0.063 & 101 & 0.99669  \\ 
  \textsc{dqn.qhi.f} & -1.895 & 1 & 1  \\ 
  \textsc{qrdqn.qhi.f} & 0.014 & 238 & 0.45904  \\ 
  \textsc{trpo.qhi.f} &  \textbf{0.046} & 345 & 0.01062  \\ 
  \textsc{a2c.qhi.f} & 0.036 & 343 & 0.01183  \\ 
  \hline
  \textsc{det.trpo.qhi} & -0.662 & 57 & 0.99985 \\ 
   \textsc{det.a2c.qhi} & -0.886 & 90 & 0.99837  \\ 
   \textsc{det.trpo.qhi.f} &  \textbf{0.032} & 271 & 0.21723  \\ 
   \textsc{det.a2c.qhi.f} & 0.02 & 287 & 0.13335  \\ 
   \hline
\end{tabular}
\caption{Comparison between the average return of each counterfactual policy and that of the \textsc{nws} policy on the evaluation years, summarized across counties (e.g. ``Median Diff." is the median difference in average return). WMW is the Wilcoxon-Mann-Whitney statistic (higher is better); its associated p-value is also included.
The first policy, marked by *, requires oracle knowledge of the future weather.}   \label{tab:main_results}
\end{table}

\subsubsection{QHI restriction is necessary for the standard RL algorithms to perform well}
Several counterfactual policies perform worse than \textsc{nws} (as indicated by negative median differences in returns): \textsc{random}, \textsc{basic.nws}, and the RL models without the QHI restriction. By contrast, the RL models with the QHI restriction perform significantly better than \textsc{nws}, as indicated by their significant positive median difference in returns. A rough estimate if \textsc{a2c.qhi} were implemented as-is across all counties in the U.S. (using the Medicare population from 2011, the midpoint of our study period) is that we could see a reduction of 222 NOHR hospitalizations per year (approximate 95\% CI = (-491, 1131); details of this CI calculation are in Appendix \ref{sec:approx-ci}). Discussion of this absolute benefit to public health is in Appendix \ref{sec:abs-health}.

\subsubsection{The futility of including future information}
Across both plain and QHI-restricted models, \textsc{trpo} was the only one that benefited from including future information in the state space. While \textsc{a2c} was minimally impacted, \textsc{qrdqn} and \textsc{dqn} were made noticeably worse. The latter may be due to the future information dramatically increasing the size of the state space/neural network parameters, which our hyperparameter tuning did not fully counteract. %
Note that for the \textsc{trpo} models, the benefit observed with the use of future information would likely be smaller in practice due to forecast / prediction error---the simulation of which is beyond the scope of this paper. Therefore, we focus on the RL models without future information in the remainder of this text.

\subsubsection{The need for stochastic policies}
Interpreting policies from the \textsc{trpo} and \textsc{a2c} models deterministically fails in the heat alerts setting under our rewards model. Additional discussion of this result is in Appendix \ref{sec:sensitivity-discussion}.
This finding raises the question: \textit{Are stochastic policies palatable from a domain perspective?}
On the one hand, it is less immediately satisfying that some aspect of heat alerts issuance would be left to chance. On the other hand, in the context of human-in-the-loop decision making---which would likely be the case for public organizations such as NWS \cite{stuart_meteo_2022}---an algorithm reporting probabilities is more informative than reporting only a binary action. In any case, if in the future a heat alerts RL model was running continuously online, it would likely need to utilize exploration (incorporating some randomness into its actions) to
update itself over time.

\subsubsection{Comparing policies that perform significantly better than NWS}
We note that the performance and behavior of \textsc{dqn.qhi} is practically identical to \textsc{aa.qhi}---not just in its numerical summary, but also in its policy behavior. 
\textsc{qrdqn.qhi} performs a bit better than \textsc{dqn.qhi}, but not as well as the stochastic policies. Between those, \textsc{a2c.qhi} performs better than \textsc{trpo.qhi}, though their policies tend to display similar characteristics. Therefore, we focus on \textsc{a2c.qhi} as the best RL policy in the post-hoc analysis.
The non-RL counterfactual policies that perform well are that of always issuing an alert on days above an optimized QHI threshold (\textsc{aa.qhi}) and of issuing alerts on the $b$ hottest days of the summer (\textsc{topk}). 
We note that the ordering of the Wilcoxon-Mann-Whitney statistics for \textsc{topk}, \textsc{aa.qhi}, and \textsc{a2c.qhi} is opposite the ordering of these policies' median difference in returns (compared to \textsc{nws}). 
This is due to the longer tails of the \textsc{a2c.qhi} differences, as illustrated in Figure \ref{fig:boxplot}. For interpreting these results, recall that \textsc{topk} is an oracle policy. However, \textsc{aa.qhi} is an implementable alternative to \textsc{a2c.qhi}, so we must consider it more seriously.

\subsection{Post-hoc Contrastive Explanation}\label{sec:post-hoc}

To characterize differences between the best RL policies, the \textsc{nws} policy, and alternative baselines, we consider both stationary features included upstream in the analysis (e.g., climate region and median household income) as well as characteristics of both the \textsc{nws} and counterfactual alert policies, namely the distributions of the days of summer on which alerts are issued and the lengths of alert streaks (sequences of repeated alerts). We start with these alert characteristics, descriptive histograms of which are in Figure \ref{fig:hist_dos}. Both \textsc{a2c.qhi} and \textsc{aa.qhi} tend to issue heat alerts earlier in the summer than \textsc{nws}. \textsc{a2c.qhi} also tends to issue shorter streaks of alerts (i.e., fewer consecutive days of alerts), which makes sense given its ability to learn sequential dependence encoded by the rewards model. 

Figure \ref{fig:boxplot} illustrates the varying performance of \textsc{topk}, \textsc{aa.qhi}, and \textsc{a2c.qhi} relative to the \textsc{nws} policy, across climate regions and counties. Note that large heterogeneity in the health effects of heat alert intervention across counties is consistent with previous findings \cite{wu_stochastic_intervention_2022}. A striking visual pattern is the increasing vertical spread from left to right: \textsc{topk} performs the least heterogeneously across counties, followed by \textsc{aa.qhi}, and finally \textsc{a2c.qhi}.
Similarly, the humid climate regions display more heterogeneity across counties than the others, across all the policies. This is partially due to those regions being larger and thus overrepresented among our 30 counties (see Table \ref{tab:counties}), but that does not fully explain the discrepancy.

To structure our investigation of why some counties see smaller (or even no) benefits from the application of the heat alerts RL, we fit CART on both the numeric difference in average return (compared to \textsc{nws}) and a polytomous indicator of which policy performed the best. 
To prevent CART overfitting to our 30 counties and to facilitate interpretation, we ensure that the minimum number of counties in each leaf (terminal) node is greater than or equal to five for the regression and three for the polytomous classification. 

\subsubsection{RL performs best in counties with larger heat alert-health signal and longer heat waves}

A classification tree distinguishing \textsc{a2c.qhi}, \textsc{aa.qhi}, \textsc{nws}, and \textsc{trpo.qhi} is shown in Figure \ref{fig:cart_nws_main}. The most predictive feature of the counterfactual policies' performance relative to \textsc{nws} is the size of the alert budget. The fact that the other policies find it harder to improve on \textsc{nws} with more alerts may be because in such settings it is less critical to discriminate days when heat alerts will be most effective, given that our rewards model assumes more alerts never hurt in an absolute sense. 
The second split in the tree indicates that the RL performs best when there is greater variation in alert effectiveness across days, in other words, when there is more signal in the heat alert-health relationship that can be leveraged by the RL. The third split highlights how it is better to use RL in counties that tend to experience more prolonged heat waves, which cause \textsc{aa.qhi} to issue more alerts in a row, increasing the likelihood of alert fatigue. 
Running the same polytomous classification on a set of more conventional covariates (Figure \ref{fig:cart_interpretable}), we see a nearly-identical tree except with median household income in place of alert effectiveness and humidity in place of prolonged heat waves. Both of these associations make sense from a domain science perspective.

\subsubsection{Additionally, RL performs better than NWS earlier in the summer}

Conducting regression on the difference in average returns between \textsc{a2c.qhi} and \textsc{nws} using CART (shown in Figure \ref{fig:cart_reg}) generated complementary insights. The most predictive feature is the median day of summer on which the RL issued heat alerts: \textsc{a2c.qhi} performs better in counties for which it identified that it is optimal to issue alerts earlier in the summer. Whereas, in counties for which it is optimal to issue alerts later in the summer, there was less room for improvement by the RL because the \textsc{nws} already tends to issue alerts later in the summer.

\section{Conclusion and Future Work Directions}\label{discussion}

This work lays the foundation of SDM for climate \& health, by (1) formulating and building an SDM pipeline to optimize heat alert issuance, via integrating traditional statistical methods with cutting-edge RL techniques, and 
(2) offering insights into scenarios where RL improves vs. fails to improve on the observed policy or simpler alternatives.

Ultimately, we found that it was necessary to restrict alerts to days with higher QHI for standard RL algorithms (\textsc{dqn}, \textsc{qrdqn}, \textsc{trpo} and \textsc{a2c}) to be effective. Our post-hoc analysis enabled investigation of where RL performed better than the \textsc{nws} policy and the simpler alternative \textsc{aa.qhi}.
Future work might explore the use of methods which ensure a new policy is never worse than the existing policy, such as safe policy learning 
or model predictive control.
Alternatively, we could develop a preliminary predictive model to select counties that are likely to benefit from RL, using intuitions similar to those in our CART analysis. 

We anticipate several common questions about our approach. The first is how to move beyond a fixed alert budget, important in a changing system (under climate change, the number of extreme heat events is expected to increase). 
From a practical standpoint, real monetary budgets from stakeholders could be used, for instance ``how many times can we afford to open our cooling center(s) this summer?" 
Future work could also explore modeling alert fatigue as part of the RL environment, for instance by incorporating tools from behavioral science. RL algorithms that are intentionally robust to distribution shift %
also merit investigation in a climate \& health setting. 

A second question is whether offline (batch) RL methods, which use only observed data in place of a simulator, could be used to circumvent dependence on the specification of a rewards model. Several major challenges stand in the way of an offline approach, highlighted in our heat alerts example. First, there is a tension between trying to improve the observed policy and controlling distribution shift \cite{levine_offline_2020}:
for instance, we see that \textsc{nws} often issues alerts in streaks, making it hard for offline RL to learn and optimize the impact of an individual alert. Second, assessing the performance of offline RL using off-policy evaluation methods is made difficult by long episodes, large potential for distribution shift (especially in the absence of a modification such as restricting alerts to very hot days), and high degree of autocorrelation in the reward \cite{uehara_ope_2022}.

We close with a broader view of climate \& health decision making. Ideally, an optimal heat alert system would account for different types of health impacts experienced by different demographic groups---future work could explore multi-objective RL. 
Similarly, it is important to recognize that the issuance of heat alerts is only a first step in reducing the public health impacts of extreme heat \cite{errett_HAPs-survey_2023}: there is much more work to be done analyzing and expanding local actions in response to heat alerts. If such on-the-ground interventions are able to increase the effectiveness of heat alerts, then this is likely to increase the ability of SDM methods such as RL to identify these effects and help us continue improving our strategies in the future.

\section*{Acknowledgements}
This work was supported by an NSF Graduate Research Fellowship (EMC), NIH award K01ES032458 (RCN), NIH award R01-ES029950 (MT, GAW), NIH award 1R01MD016054-01A1 (FD), NIH award 5R01ES030616-04 (MT), NIH award 3RF1AG074372-01A1S1 (MT), NIH award P30ES000002 (MT), Alfred P. Sloan Foundation grant G-2020-13946 (FD), the Fernholz Foundation (MT), and Wellcome Trust grant 216033-Z-19-Z (GAW). 
We thank the National Studies on Air Pollution and Health research group for their support of this project. The computation in this paper was performed on the FASSE and FASRC Cannon clusters supported by the FAS Division of Science Research Computing at Harvard University. Lastly, we thank Kate Weinberger for facilitating access to the heat alerts data set, and the STAT 234 teaching staff at Harvard University for their feedback during early stages of this project. 

\emph{Disclosures}: GAW currently serves as a consultant for the Health Effects Institute (Boston, MA) and recently served as a consultant for Google, LLC (Mountain View, CA).

\clearpage
\bibliography{references}

\newcommand{\mysection}[1]{%
  \refstepcounter{section} %
  \renewcommand{\thesection}{\Alph{section}} %
  \section*{\thesection\hspace{1em} #1} %
}

\newcommand{\mysubsection}[1]{%
  \refstepcounter{subsection} %
  \renewcommand{\thesubsection}{\Alph{section}.\arabic{subsection}} %
  \subsection*{\thesubsection\hspace{1em} #1} %
}

\appendix

\renewcommand{\thefigure}{S\arabic{figure}}
\renewcommand{\thetable}{S\arabic{table}}

\renewcommand{\thesection}{\Alph{section}}

\mysection{Heat Alerts MDP Terminology}\label{app:terms}

Table \ref{tab:terms} summarizes the heat alerts MDP components.

\begin{table*}[ht]
\small
  \centering
  \scalebox{0.95}{
    \begin{tabular}{ p{3cm} | p{3cm} | p{10cm} }
      \toprule
      \textbf{RL Term} & \textbf{Notation} & \textbf{Meaning in the Heat Alerts Application} \\
      \midrule
      episode & $t \in \{0,\ldots, H - 1\}$ & the $H=152$ days of the warm season/summer in a specific year and county; $H$ is the horizon
      \\
      \hline
      state & $s_t=(\xi_t, x_t)$ & a vector of time-varying factors influencing the
      effect of heat on hospitalizations and the effectiveness of an alert; $\xi_t$ is the exogenous component (e.g., heat index) and $x_t$ is the endogenous component (e.g., history of heat alerts)
      \\
      \hline
      action & $a_t$ & whether a heat alert is issued (1) or not (0) at time $t$\\
      \hline
      reward & $r_t=R(s_t, a_t)$ & daily rate of heat-related hospitalizations at time $t$; a negative sign transform is applied so that fewer hospitalizations corresponds to a greater reward \\
      \hline
      agent & --- & the algorithmic decision-maker that implements a policy to determine when to issue heat alerts \\ 
      \hline
      policy & $\pi(a_t | s_t)$ & the heat alert issuance rule;
      the NWS policy is the rule that was used to issue heat alerts in the U.S. from 2006-2016
      \\
      \hline
      cumulative \mbox{reward / return} & $\sum_{h=t}^{H - 1} r_h$ & sum of remaining reward (transformed hospitalizations) at time $t$; unless otherwise specified, ``cumulative" implies $t=0$
      \\
      \hline
      \mbox{intervention /} alert budget & $\sum_{t=0}^{H-1}a_t \leq b$ & the maximum number of heat alerts that can be issued in a given county-summer; in our experiments, the observed number of NWS alerts \\
      \bottomrule
    \end{tabular}
    }
    \caption{MDP/RL terminology in the context of the heat alerts environment.} 
    \label{tab:terms}
  \end{table*}

\mysection{Details of the Data Set}\label{sec:data-si}

Evidence from the climate \& health and environmental epidemiology literature guides our selection of relevant variables as well as feature engineering. 

\mysubsection{Heat Index and Heat Alerts} We follow \citet{weinberger_heat_2021} and \citet{wu_stochastic_intervention_2022} in using population-weighted maximum heat index to describe heat exposure in each county. They also combined NWS heat ``warnings" and ``advisories" into the single category of ``alerts". In some locations, where the forecast zones in which the NWS issues alerts are smaller than county boundaries, the alert history was taken from whichever forecast zone contained the largest proportion of the 2010 county population.

\mysubsection{Medicare Hospitalizations} Medicare is available for people aged 65 years and older. The hospitalization claims data we use in this study is from Medicare Part A, fee-for-service. A 2021 report \citeA{medicarereport} found that fee-for-service enrollees do not significantly differ from managed care enrollees (the rest of the Medicare population). 

\begin{figure}[ht]
  \centering
  \includegraphics[width=0.45\textwidth]{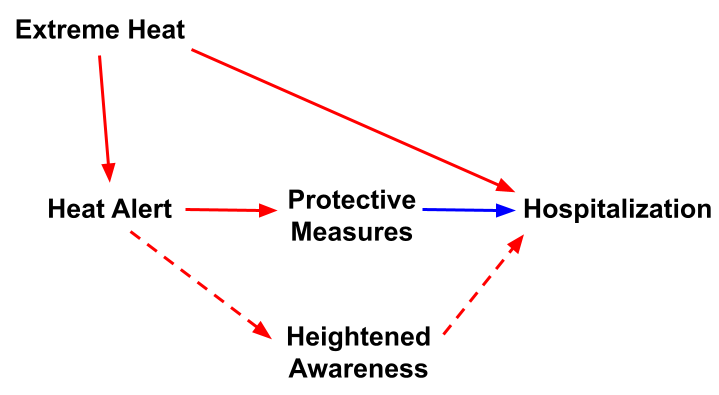}
  \caption{Unobserved mediation of the effect of heat alerts on hospitalizations by heightened awareness.
  } 
  \label{fig:mediation}
\end{figure}

Figure \ref{fig:mediation} illustrates the rationale for excluding obviously heat-related hospitalization types from the pooled outcome, as described in Section \ref{sec:data}.

\mysubsection{Descriptive Statistics of Heat, Alerts, and Hospitalizations}\label{sec:descriptive}

\begin{table*}[ht]
\small
\begin{center}
\begin{tabular}{ p{4cm} p{1cm}p{1cm}p{1cm}p{1cm}p{1cm}p{1cm} } 
  \hline
  \textbf{Variable} & \textbf{Min.} & \textbf{Q1} & \textbf{Median} & \textbf{Mean} & \textbf{Q3} & \textbf{Max.} \\
  \hline
   *NOHR hosps per 1,000 %
   & 5.4 (3.1) & 12.5 (12.7) & 14.8 (15.2) & 15 (15.3) & 17.3 (17.8) & 25.7 (34.5) \\
    *No. of NWS alerts & 0 (0) & 4 (0) & 8 (3) & 9.6 (4.8) & 13 (7) & 43 (57) \\ 
    DMHI on alert days (\textdegree F) & 70.5 (68.3) & 100.3 (99.7) & 103.5 (103.2) & 102.5 (102.6) & 106.2 (105.9) & 121.1 (144) \\ 
    Q-DMHI on alert days & 18.9 (15.6) & 91.6 (92.6) & 95.8 (96.9) & 93.4 (94.2) & 98.2 (98.9) & 100 (100) \\
    *No. of 90th pct. HI days &  0 (0) & 9 (8) & 14 (14) & 15.4 (15.4) & 21 (21) & 46 (61) \\
  DOS of NWS alerts 
   &  16 (2) & 68 (69) & 84 (84) & 83.4 (83.5) & 99 (99) & 142 (150) \\
   DOS of 90th pct. HI days &  15 (1) & 72 (69) & 88 (85) & 86.9 (85.2) & 102 (102) & 145 (153) \\
  \hline
\end{tabular}
\caption{Summary statistics for our 30 counties of interest (and all the counties) %
of Medicare NOHR hospitalizations, NWS heat alerts, heat index, and day of summer for several features of interest. DMHI = daily maximum heat index, Q-DMHI = county quantile of DMHI (referred to simply as QHI in the main text), DOS = day of summer (out of 152). Variables given per county-summer %
are marked with *.} \label{descriptive}
\end{center}
\end{table*}

Table \ref{descriptive} provides descriptive statistics of the heat alerts dataset, both for the 30 counties in our RL analysis and all the counties with population $\geq$ 65,000 (used to create the BROACH environment). The two sets of statistics are similar aside from the number of heat alerts issued because the counties in the RL analysis were selected specifically for having issued higher numbers of alerts. Other observations are the long tail of the number of NWS alerts per county-summer, that some heat alerts were issued on days that were not very hot in an absolute or relative (quantile) sense, that there were some county-summers with few ``hot" days (90th percentile by county\footnote{Here, the 90th percentile was chosen to provide some intuition for the distribution of heat index and heat alerts. For context, Hondula et al. (2022) found that 95th percentile daily maximum heat index was the climatological indicator most strongly correlated with NWS heat alert frequency.}) and others with many, and that the distribution of day-of-summer for NWS heat alerts and 90th percentile QHI by county are quite similar. %
A few other relevant statistics not shown in the table are that (across all the counties) there were a total of 40,387 heat alerts issued between 2006 and 2016, that 17.2\% of NWS heat alerts were issued on days with heat index values that were \textit{not} in their county's top 90th percentile, and that 74.0\% of days in the top 90th QHI by county did not have NWS heat alerts.

\mysection{Details of the Bayesian Rewards Model}\label{sec:r-model-details}

Recall our formulation for the expected per capita rate of NOHR hospitalizations at time $t$ in county-summer $(k,j)$ when taking action $a$:

$$
\rho_t^{(k,j)}(a) =  \lambda_k(s_t^{(k,j)}) (1 - a\cdot\tau_k(s_t^{(k,j)}))
$$

\paragraph{The baseline and effectiveness functions}  Note that $\lambda_k$ and $\tau_k$ are unique to each location $k$, allowing for spatial heterogeneity and thus reducing the possibility of unobserved spatial confounding \citeA{bell2015explaining, nobre2021effects, tec2024space}.  However, this design introduces the challenge of estimating these functions under limited data for each county (11 summers between 2006 and 2016). This challenge is amplified by the low signal in the health effects of heat and heat alerts \cite{weinberger_heat_2021}. To address this, we adopt a (generalized) linear model for $\lambda_k$ and $\tau_k$ paired with (i) borrowing statistical strength across counties using a data-driven random effects prior and (ii) injecting domain knowledge on the signs of certain coefficients. The functions $\lambda_k$ and $\tau_k$ are specified as
\begin{equation}
  \label{eq:lambda-tau}
  \begin{aligned}
      \lambda(s_t^{(k,j)}) &:= \exp\left(\beta_k^\top s_t^{(k,j)}\right), \\
      \tau(s_t^{(k,j)}) &: = \text{sigmoid}\left(\delta_k^\top s_t^{(k,j)}\right),
  \end{aligned}
  \end{equation}
where $\beta_k$ and $\delta_k$ are the model coefficients for each county $k\in\gK$. While this specification is a linear combination of state variables, we allow for non-linear effects by using a piecewise linear basis expansion on the quantile of heat index (QHI) and a spline basis on the day of summer. 
The piecewise linear expansion on QHI in $\lambda$ (the baseline function) has knots at the 25th and 75th percentiles. In $\tau$ (the alert effectiveness function), we include only a linear term for QHI---again, conditional on day of summer and excess QHI (compared to the average of the last 3 days, a time window informed by the literature on heat waves and health \citeA{metzger_heat_waves_2010})---to both stabilize effect identification and facilitate interpretation. The spline on day of summer (with 3 degrees of freedom) is included in both $\lambda$ and $\tau$.

\begin{figure*}[ht]
\begin{center}
\includegraphics[width=0.65\textwidth]{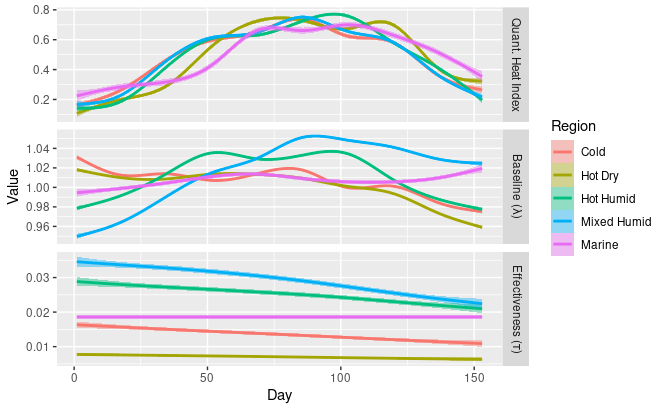}
\end{center}
\caption{Smoothed trajectories of observed quantile of heat index, modeled baseline NOHR hospitalization rate, and modeled alert effectiveness (assuming no past alerts) across days of summer; colored by climate region. Note that $\lambda$ and $\tau$ are normalized as they are in the RL reward function.} \label{fig:time_series}
\end{figure*}

Figure \ref{fig:time_series} illustrates the interaction of $\lambda_k$ and $\tau_k$ to capture the total effect of heat alerts on NOHR hospitalizations. For instance, we see that in early summer it is possible to have high heat alert effectiveness even when the heat index is not as high as it is later in the summer, possibly because people are less well-adapted to heat and/or more responsive to heat alerts early on. 

\paragraph{Data-driven prior} We propose a data-driven prior distribution that allows borrowing statistical strength from other locations while preserving the possibility of spatial heterogeneity. Let $w_k$ denote the vector of county-level features that are not included in the state vector but are predictive of the effect of heat and heat alerts on hospitalizations (see ``County-level characteristics'' in \ref{sec:data}), and let $\gamma_k = [\beta_k; \delta_k]=(\gamma_k^\ell)_{\ell=1}^L$ denote the parameters corresponding to county $k$. Let N() denote a normal distribution. Then, the prior distribution of each $\gamma_k^\ell$ is set independently as $p_\theta(\gamma_k^\ell | \sigma_\ell, w_k)$ =

\begin{equation}
  \label{eq:prior-beta}
  \begin{aligned}
   \begin{cases}
\mathrm{N}(\gamma_k^\ell; f_{\theta}(w_k)_\ell, \sigma_\ell^2) & \text{ no domain knowledge}, \\
\mathrm{LogN}(\gamma_k^\ell; \exp(f_{\theta}(w_k)_\ell), \sigma_\ell^2) & \text{ if }
\gamma_k^\ell \in (0, \infty), \\
\mathrm{NegLogN}(\gamma_k^\ell; -\exp(f_{\theta}(w_k)_\ell), \sigma_\ell^2) & \text{ if } 
\gamma_k^\ell \in (-\infty, 0) \\
\end{cases} && \\
\end{aligned}
\end{equation}
where $f_{\theta}$ is a feed-forward neural network (with weights $\theta$ and $L$ outputs) specifying the prior mean as a function of the spatial features, and $\sigma_\ell$ specifies the prior scale. We use a standard hyper-prior $p(\sigma_\ell)= \mathrm{HalfCauchy(\sigma_\ell)}$ \citeA{gelman2006prior}.

The domain knowledge constraints we utilize are: (i) past heat alerts cannot increase the baseline hospitalization rate---in other words, alert fatigue can be identified only in $\tau$---and (ii) higher QHI cannot decrease the effectiveness of heat alerts---note that this is conditional on day of summer. 

The weights $\theta$ are learned from the data to characterize the prior distribution that best describes the latent space according to the data. As remarked in \citeA{Wang_Miller_Blei_2019}, this procedure can be seen as a form of Empirical Bayes. According to this viewpoint, we can interpret the prior mean function $f_{\theta}$ in \eqref{eq:prior-beta} as an encoder that maps the county-level features $w_k$ to a data-driven probability distribution over a latent space. The coefficients $\gamma_k$ are latent variables in this space. The predictive model $\rho_t^{(k,j)}(a)$ in \eqref{eq:hospitalization-model} acts as a decoder. 

\paragraph{Training} Traditional Bayesian inference methods using Markov Chain Monte Carlo are computationally prohibitive for our model due to the high dimensionality of parameters
and the number of observations for each county.
To address this issue, we use variational inference, which introduces an approximation of the true posterior distribution of the model parameters. We summarize the approach. 

Let $(\gamma, \sigma) \sim q_{\psi}$ denote an approximation of the true posterior distribution where $\psi$ are the learnable parameters of the approximating distribution $q_\psi$. Next, let $o_t^{(k,j)}=(s_t^{(k,j)}, a_t^{(k,j)}, y_t^{(k,j)}, n^{( k, j)})$ denote a data point in the training set and $\gL(o_t^{(k,j)}; \gamma_k)$ be the Poisson log-likelihood corresponding to this data point (see \eqref{eq:hospitalization-model}). Variational inference attempts to maximize the likelihood of the observed data while penalizing the divergence between the variational posterior $q_\psi$ and the prior $p_\theta(\gamma, \sigma | w)= {\prod_{k,l}} \, p_\theta(\gamma_k^\ell | \sigma_\ell, w_k)\, {\prod_{l}}\, p(\sigma_\ell)$. The optimization objective is to maximize the evidence lower bound (ELBO) given by
\begin{equation}
\label{eq:elbo}
\begin{aligned}
\textrm{ELBO}(\psi, \theta) &=  \E_{q_{\psi}}\Big[{\textstyle{\sum_{k, j, t}}}\gL(o_t^{(k,j)}; \gamma_k)\Big] \\ &\:\:\:\:\:\:\:\: - \KL\left(q_{\psi}(\gamma, \sigma) \,\Vert\, p_\theta(\gamma, \sigma | w)\right), \\
\end{aligned}
\end{equation}
where $\KL$ denotes the Kullback-Leibler divergence. 
To ensure the variational posterior is flexible enough to approximate the true posterior, we use a low-rank multivariate normal distribution \citeA{tomczak_low-rank_2020}. This approach uses a bijective transform of the parameters to an unconstrained space and then applies a low-rank factorization to the covariance matrix of the multivariate normal distribution. Thereby, it allows for efficient computation while approximating the true correlation structure of the joint posterior distribution. The data-driven prior parameters $\theta$ are learned jointly with the variational parameters $\psi$ using stochastic optimization. Specifically, we train one neural network for $\beta$ (the coefficients of the baseline function) and another for $\delta$ (the coefficients of the alert effectiveness function). Both networks have one hidden layer with 32 units---this modest architecture avoids overfitting the relatively small number of spatial features in our dataset. 

\begin{figure}[ht]
  \begin{center}
    \includegraphics[width=0.35\textwidth, trim={0 0 0 1.1cm},clip]{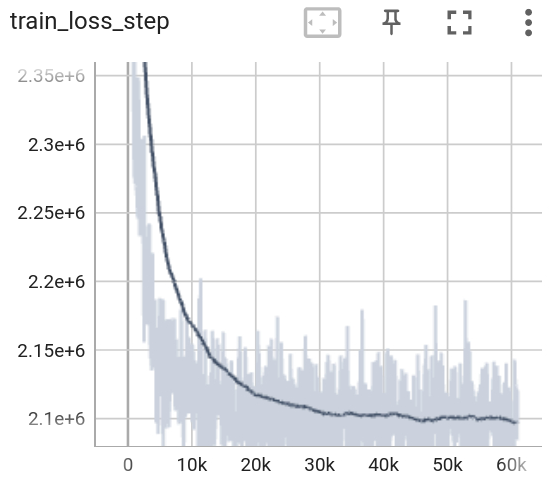}
  \end{center}
  \caption{Convergence of the ELBO in the Bayesian rewards model during training (80 epochs).} \label{fig:bayesian_loss}
  \end{figure}

\paragraph{Results} The rewards model clearly converged in its loss across epochs (shown in Figure \ref{fig:bayesian_loss})
and achieved higher predictive accuracy than standard machine learning algorithms (i.e., random forest and neural network). Predicting the rate of NOHR hospitalizations in the Medicare population in each county, this model has $R^2 = 0.102$---using the calculation $SSR/SST$ for comparability with earlier machine learning models (such as that we used when checking the signal in the mortality data). Predicting the absolute number of NOHR hospitalizations, it has $R^2 = 0.884$---much higher because the value is mostly driven by county population size.
The model also displayed very good coverage when we ran it on synthetic data (1,000 samples from the posterior predictive) using known coefficients: average coverage across parameters for a 90\% CI was 0.897. 

\begin{figure}[ht]
  \begin{center}
  \includegraphics[width=0.49\textwidth]{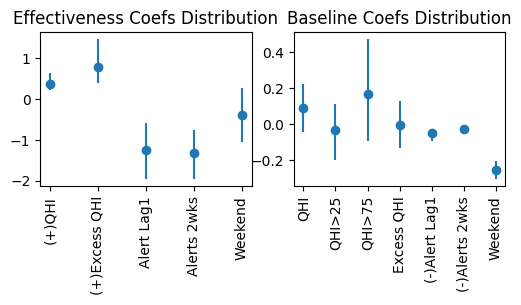}
  \end{center}
  \caption{A sample of posterior coefficients from the rewards model (in $\tau$ and $\lambda$ respectively), displaying IQR and median. Bias and coefficients for the spline on day of summer are not shown, for ease of interpretability. (+)/(-) indicates a coefficient constrained to be positive/negative.} 
  \label{fig:bayes_coefs}
  \end{figure}

Figure \ref{fig:bayes_coefs} shows one posterior sample across the 761 counties on which the model was fit. Note that the spread across counties is larger than the spread within counties, so the IQR bars are no larger for multiple samples. Overall, we see that the coefficients on past alerts in today's alert effectiveness ($\tau$) are both negative, empirical evidence of alert fatigue. In the baseline hospitalization rate ($\lambda$), we see a strong protective effect by weekend.

\mysection{Details of the RL Analysis}\label{app:rl-details}

\mysubsection{Selection of 30 Counties}\label{select-30-counties}

\begin{table*}[ht]
\small
\centering
\begin{tabular}{ p{2cm}p{1cm}p{1.5cm}p{1.5cm}p{1.5cm}p{1.5cm}p{1.5cm}p{1cm}}
  \hline
\mbox{County Fips} (\mbox{State name} \mbox{abbreviation)} & Region & \mbox{No. Alerts} (Eval Years) & Population Density (/mi$^2$) & \mbox{Median HH} \mbox{Income} (USD) & Democratic Voters (\%) & Broadband Usage (\%) & Average PM$_{2.5}$ ($\mu g/m^3$)\\ 
  \hline
5045 (AR) & MxHd &  61 & 179.00 & 50314.00 & 34.00 & 37.50 & 9.08 \\ 
  20161 (KS) & MxHd &  47 & 120.00 & 43962.00 & 43.50 & 52.20 & 6.35 \\ 
  21059 (KY) & MxHd &  24 & 212.00 & 46555.00 & 38.70 & 53.30 & 10.22 \\ 
  29019 (MO) & MxHd &  50 & 242.00 & 48627.00 & 51.60 & 46.60 & 7.85 \\ 
  37085 (NC) & MxHd &  31 & 200.00 & 44625.00 & 39.30 & 46.70 & 9.65 \\ 
  40017 (OK) & MxHd &  63 & 133.00 & 63629.00 & 22.80 & 70.10 & 7.53 \\ 
  47113 (TN) & MxHd &  58 & 176.00 & 41617.00 & 44.30 & 79.40 & 9.29 \\ 
  42017 (PA) & MxHd &  25 & 1036.00 & 76555.00 & 51.00 & 85.10 & 9.14 \\ 
  41053 (OR) & Mrn &  31 & 102.00 & 52808.00 & 45.60 & 79.20 & 3.24 \\ 
  41067 (OR) & Mrn &  31 & 745.00 & 64180.00 & 57.70 & 82.70 & 4.28 \\ 
  53015 (WA) & Mrn &  23 & 90.00 & 47596.00 & 49.40 & 53.60 & 3.59 \\ 
  13031 (GA) & HtHd &  43 & 106.00 & 35840.00 & 38.90 & 57.50 & 9.01 \\ 
  22109 (LA) & HtHd &  34 & 91.00 & 49960.00 & 27.90 & 55.50 & 8.26 \\ 
  28049 (MS) & HtHd &  57 & 283.00 & 37626.00 & 69.90 & 36.50 & 8.50 \\ 
  45015 (SC) & HtHd &  48 & 168.00 & 52427.00 & 41.20 & 66.20 & 8.72 \\ 
  48157 (TX) & HtHd &  42 & 707.00 & 85297.00 & 47.80 & 83.70 & 8.64 \\ 
  48367 (TX) & HtHd &  41 & 131.00 & 64515.00 & 18.30 & 52.20 & 7.84 \\ 
  22063 (LA) & HtHd &  42 & 201.00 & 56811.00 & 13.70 & 47.60 & 8.87 \\ 
  4015 (AZ) & HtDr &  29 & 15.00 & 39200.00 & 28.70 & 63.60 & 4.59 \\ 
  6025 (CA) & HtDr &  40 & 42.00 & 41807.00 & 64.00 & 64.40 & 7.22 \\ 
  32003 (NV) & HtDr &  29 & 251.00 & 52873.00 & 55.90 & 78.10 & 5.21 \\ 
  4013 (AZ) & HtDr &  72 & 423.00 & 53596.00 & 44.00 & 77.40 & 6.38 \\ 
  6071 (CA) & HtDr &  29 & 103.00 & 54090.00 & 51.60 & 77.10 & 6.38 \\ 
  17115 (IL) & Cold &  22 & 190.00 & 46559.00 & 45.70 & 63.60 & 8.87 \\ 
  17167 (IL) & Cold &  25 & 228.00 & 55449.00 & 46.00 & 66.50 & 8.75 \\ 
  19153 (IA) & Cold &  24 & 764.00 & 59018.00 & 54.90 & 57.90 & 8.25 \\ 
  19155 (IA) & Cold &  27 & 98.00 & 51304.00 & 44.30 & 42.00 & 7.89 \\ 
  34021 (NJ) & Cold &  29 & 1639.00 & 73480.00 & 67.30 & 81.60 & 9.31 \\ 
  29021 (MO) & Cold &  44 & 219.00 & 44363.00 & 43.70 & 48.80 & 7.97 \\ 
  31153 (NE) & Cold &  27 & 681.00 & 69965.00 & 37.60 & 72.60 & 7.92 \\ 
   \hline
\end{tabular}
\caption{Descriptive statistics of the 30 counties used in policy optimization. Climate region abbreviations = \{MxHD: Mixed-Humid, Mrn: Marine, HtHd: Hot-Humid, HtDr: Hot-Dry, Cold: Cold\}. 
HH = household. PM$_{2.5}$ is an annual average.}\label{tab:counties}
\end{table*}

Out of the 170 counties with 75 or more heat alerts issued between 2066 and 2016, we selected 30 using the following criteria: for each climate region\footnote{Excluding ``Mixed Dry" and ``Very Cold" because there were very few counties in these regions with sufficient population size}, select the county with the highest variance in estimated heat alert effectiveness (indicating higher signal as identified by our rewards model). To get up to 30, repeat this selection, but first prioritize selecting a county from a U.S. state not already in the set, and if we must repeat a state, do not select a county that is directly adjacent to one already in the set. If there are no more eligible counties in a region, select from a larger region. Characteristics of these 30 counties are provided in Table \ref{tab:counties}.

\mysubsection{RL Algorithm Descriptions}\label{rl-algos}

\textsc{dqn} is standard Deep Q-learning. \textsc{qrdqn} differs from \textsc{dqn} in that rather than estimating the average return from taking action $a$ at state $s$, it estimates the distribution over the returns using quantile regression. 
Compared to other common stochastic policy gradient algorithms, \textsc{trpo} is distinguished by its constraint on the size of the optimization step, so that the new policy lies within a ``trust region” from the previous policy, in which local approximations of the policy function are still accurate \citeA{han_2023}.  
\textsc{a2c} improves on the basic setup of actor-critic by updating its policy model using the advantage function, which indicates how much better an action is than the alternative(s), rather than just the state-action value of that one action; this helps increase model stability \citeA{han_2023}.

\mysubsection{Hyperparameter Tuning}\label{tuning}

RL algorithms are known to be sensitive to their hyperparameters, especially in empirical settings \citeA{eimer_hyperparameters_2022}. %
To address this concern, we tuned hyperparameters per county, using the average return metric. First, we did a broad sweep on one county from each of the five climate regions using \textsc{trpo.qhi}, experimenting with values of learning rate $\in \{0.01, 0.001, 0.0001\}$, discount factor $\in \{1.0, 0.999, 0.99\}$, neural network architecture (number of hidden layers $\in \{1, 2, 3\}$ and number of hidden units $\in \{16, 32, 64\}$), and size of the replay buffer $\in \{1024, 2048, 4096\}$; paired with possible QHI thresholds $\in \{0.55, 0.7, 0.85\}$ and both including and not including future information. After determining that learning rate=0.001 and discount factor=1.0 performed the best across the board, we tuned the remaining parameters for each county and each RL model (all combinations of \textsc{trpo}, \textsc{a2c}, \textsc{qrdqn} and \textsc{dqn} with and without QHI restrictions and future information). To make the computation time more reasonable, we did a grid search on two values each for the number of hidden layers (2 or 3), number of hidden units (16 or 32), and size of the replay buffer (1,500 or 3,000). 
For \textsc{qrdqn} specifically, we also tested different values for the number of quantiles, which is closely tied to computation time for this algorithm. Using 10 quantiles rather than 20 ended up being not only faster but also better, potentially due to the latter overfitting. 

\mysubsection{Sensitivity Analysis Using 2014-2016 as the Evaluation Years}\label{seq-eval}

For real-world deployment of a heat alert-RL system, an ideal setup would be to train on years $\{1,...,x\}$ and evaluate on year $x+1$. However, in this analysis we held out $\{2007, 2011, 2015\}$ as the evaluation set to provide a better representation of historical performance. Note that while days within each episode are consecutive, the 7-month gap between episodes lessens concern about correlated episodes. 

\begin{table}[!ht]
\centering
\begin{tabular}{p{2cm}p{1.2cm}p{1.5cm}p{1.2cm}}
\hline
\textbf{Evaluation Period} & \textbf{QHI Median} & \textbf{Median Diff. (RL-NWS)} & \textbf{WMW p-value} \\ \hline
\multicolumn{4}{c}{Main Analysis:}\\ \hline
\{2007, 2011, 2015\} &  & 0.042 & 0.011 \\ 
2007             & 0.51   & 0.006 & 0.013 \\ 
2011             & 0.555  & 0.016 & 0.027 \\ 
2015             & 0.519  & 0.01  & 0.010 \\ \hline
\multicolumn{4}{c}{Sensitivity Analysis:}\\ \hline
\{2014, 2015, 2016\}      &        & 0.007 & 0.121 \\ 
2014           & 0.46   & 0.002 & 0.012 \\ 
2015           & 0.519  & 0.009 & 0.022 \\
2016           & 0.6    & -0.005 & 0.777 \\ \hline
\end{tabular}
\caption{Comparison of results using 2014-2016 as the evaluation years (sensitivity analysis) as opposed to 2007,2011,2015 (main analysis).}
\label{tab:seq-eval}
\end{table}

To provide intuition for how sequential training-testing would look, we re-ran the best RL model (\textsc{A2C.QHI}) using 2006-2013 to train and 2014-2016 to evaluate. To reduce the computational burden, we used the optimized hyperparameters from the main analysis for each county. Table \ref{tab:seq-eval} shows the aggregated results as well as results stratified by year. A few notable results are that (i) the ordering of the original results ($2007 < 2015 < 2011$) is the same as the ordering of median QHI, (ii) the 2015 results are similar in each evaluation scheme, and (iii) the 2016 result is the worst. However, this last result is overly bleak because in an online deployment we would not be evaluating 3 years ahead---the model would be updated at least once a year.

\mysection{Summary of the Workflow Using BROACH}\label{viz-summary} 

This section is meant to illustrate a toy example of our workflow. Imagine seven counties located in two climate regions, as shown in the figure below. 
\begin{figure}[ht]
\begin{center}
\includegraphics[width=0.49\textwidth, trim={4cm, 14cm, 4cm, 1.9cm}, clip]{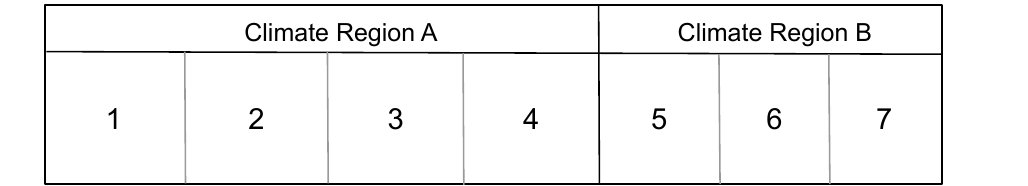} %
\end{center}
\end{figure}

Let the training years TY = $\{2006, 2008, 2009, 2010, 2012, 2013, 2014, 2016\}$ and evaluation years EY = $\{2007, 2011, 2015\}$.

First, train the rewards model on counties 1-7 and years TY $\cup$ EY to obtain posterior distributions for $r_1(s,a), \: ... , r_7(s,a)$. Then, for each county, follow the steps below. Consider county 1 for illustration. Our procedure is as follows:

\begin{algorithmic}[1]
\If{Training RL (with a QHI restriction)} 
    \For{h in $\{0.5, 0.55, \: ..., \: 0.9\}$}
        \For{i in 1 to $N_T$ episodes}
            \State Sample $r_{1,i}(s,a)$ from the posterior for $r_1(s,a)$
            \State Sample $\xi_i \sim TY \times \{\xi_1, \xi_2, \xi_3, \xi_4 \}$
            \State Update $\pi_{1,h}(a|s)$ s.t. a = 0 if QHI $<$ h, using $r_{1,i}$ and $\xi_i$
        \EndFor
        \State Select $\pi^*_{1,h}$ (across i) with the largest average return using $\xi_i \sim EY \times \{\xi_2, \xi_3, \xi_4 \}$ 
    \EndFor 
\ElsIf{Tuning RL or \textsc{aa.qhi}}
    \For{h in $\{0.5, 0.55, \: ..., \: 0.9\}$}
        \For{i in 1 to $N_E$ episodes}
            \State Sample $r_{1,i}(s,a)$ from the posterior for $r_1(s,a)$
            \State Sample $\xi_i \sim EY \times \{\xi_2, \xi_3, \xi_4 \}$
            \State Calculate cumulative reward using $\pi^*_{1,h}$ (or \textsc{aa.qhi} with h), $r_{1,i}$ and $\xi_i$
        \EndFor
    \EndFor 
    \State Select h$^*$ with the largest return
\ElsIf{Final evaluation for any policy (but using notation for optimized RL)}
    \For{i in 1 to $N_E$ episodes}
        \State Sample $r_{1,i}(s,a)$ from the posterior for $r_1(s,a)$
        \State Sample $\xi_i \sim EY \times \{\xi_1\}$
        \State Calculate cumulative reward using $\pi^*_{1,h^*}$, $r_{1,i}$ and $\xi_i$
    \EndFor
\EndIf
\end{algorithmic}

\mysection{Approximate Confidence Interval Calculation}\label{sec:approx-ci}

To generate an approximate confidence interval for the absolute number of NOHR hospitalizations saved by implementing our best RL model, we use the estimates from the final evaluation for \textsc{a2c.qhi} and \textsc{nws}. However, rather than first calculating average returns per county (across the 1,000 evaluation episodes) and then taking the median across the 30 counties, we first take the median return (across the 30 counties) from each of the 1,000 evaluation episodes and then calculate the 2.5\% and 97.5\% quantiles of the resulting distribution of medians.  

\mysection{Discussion of Absolute Health Benefits}\label{sec:abs-health}

Do we estimate substantive health benefits from implementing the improved heat alerts policies detailed in the main text? 
For general context, the median rate of NOHR hospitalizations per 10,000 Medicare enrollees per summer during our study period was 148 (Table \ref{descriptive}), and we estimated that deploying our best RL model \textit{relative to issuing no heat alerts} would save 2 of these 148, or 1.4\%, in the 30 counties in our RL analysis. Noting that the overall relative risk of a Medicare enrollee experiencing one of the NOHR hospitalization types on a heat wave day is about 1.10 \cite{bobb_2014}, this is substantial evidence in support of continuing to issue heat alerts to protect public health. %

Compared to the observed \textsc{nws} policy, \textsc{a2c.qhi} applied to the 30 counties in our analysis would save approximately 0.042 NOHR hospitalizations per 10,000 Medicare enrollees per summer, a reduction of only 0.03\%. Multiplying this by 49 million, the size of the Medicare population in 2011 (the midpoint of our study period), yields about 222 NOHR hospitalizations saved per summer. However, this modest value should be further contextualized by several points.

Firstly, heat alerts by themselves (and in our case, the re-arrangement of the same number of heat alerts) are a very cost-effective intervention, so even small reductions in health harms are promising. 

\begin{figure*}[ht] 
\begin{center}
\includegraphics[width=0.7\textwidth]{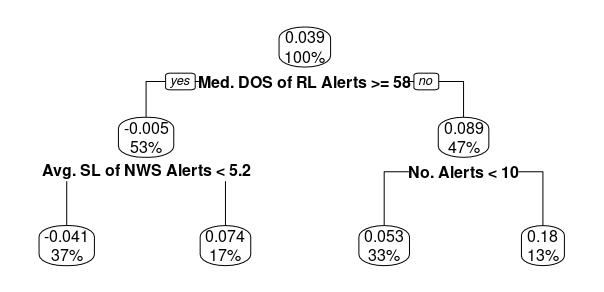}
\end{center}
\caption{Regression tree on the differences in average return between \textsc{a2c.qhi} and \textsc{nws}. Within each node, the number at the top is the prediction (average) difference, with positive indicating that the RL is better and negative indicating that the \textsc{nws} is better. The percentage at the bottom is the fraction of counties represented by that node. DOS is the day of summer, SL is the streak length, and ``No. Alerts" is the average number of heat alerts per warm season during the evaluation years.} \label{fig:cart_reg}
\end{figure*}

Secondly, if we take into account the large heterogeneity in the benefit of RL across counties with numbers from the CART regression (Figure \ref{fig:cart_reg}), assuming that we figure out how to implement a safe policy such that counties which would not benefit are unaffected, this number increases to 262 ((49m/10k)*(0.074*0.17 + 0.053*0.33 + 0.18*0.13)) NOHR hospitalizations saved per summer. We obtain a nearly identical result from running a CART analysis using only variables which are available for all 761 counties that were included in the rewards model, and applying the results of that model across the whole set of counties. These are rough approximations given that they assume the proportion of counties in which RL alert policies were found to improve health outcomes in our sample is generalizable to the rest of the US. It is plausible that our sample of 30 counties is unrepresentative of the U.S. as they tend to have higher variability in alert effectiveness (increasing potential improvements from RL) and also higher alert budgets. 

Thirdly, these metrics reflect the climate and Medicare population between 2006 and 2016. We confidently anticipate that both the frequency of extreme heat events under climate change and the size of the Medicare population will continue increasing \citeA{dahl_increased_heat_2019}. For instance, while there were 49 million Medicare enrollees in 2011, by 2023 this number has increased to over 65 million, and by 2050 it is projected to be over 85 million \citeA{cms_trustees_2023}.

Fourthly, the numbers reported in this study should not be interpreted as the overall benefit to public health from improving heat alerts issuance strategies. The health data used in our study only reflect inpatient hospitalizations from NOHR causes billed to Medicare Part A, or hospital fee-for-service. Only about 60\% of Medicare enrollees are in the fee-for-service population\footnote{\url{https://www.cms.gov/Research-Statistics-Data-and-Systems/Statistics-Trends-and-Reports/Beneficiary-Snapshot/Downloads/Bene_Snaphot.pdf}}, so hospitalizations of the other 40\% do not appear in our dataset.

\mysection{Extended Results from the RL Analysis}

\mysubsection{Policy Examples}

\begin{figure}[ht]
\begin{center}
\includegraphics[width=0.45\textwidth]{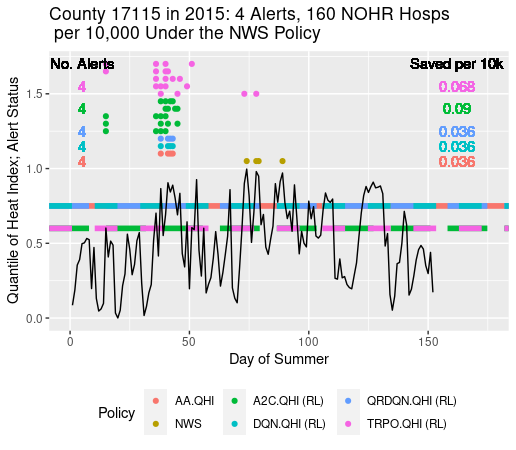}
\end{center}
\caption{Example of a setting with relatively few heat alerts.} 
\label{fig:mini-case study - few alerts}
\end{figure}

\begin{figure}[ht]
\begin{center}
\includegraphics[width=0.45\textwidth]{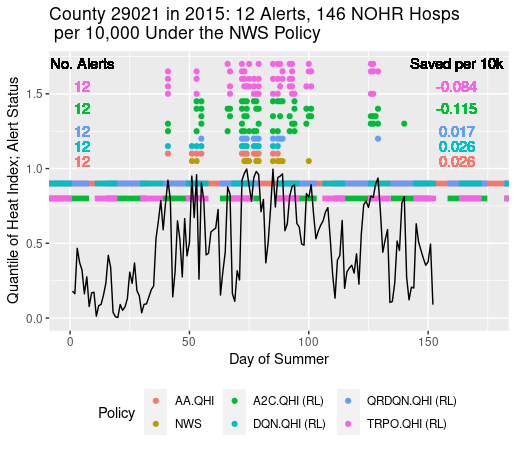}
\end{center}
\caption{Example of a setting where implementing RL would not be beneficial relative to \textsc{nws} or \textsc{aa.qhi}.} 
\label{fig:mini-case study - poor RL}
\end{figure}

\mysubsection{Additional Discussion of Stochastic vs Deterministic Policies}\label{sec:sensitivity-discussion}

Interpreting policies from the \textsc{trpo} and \textsc{a2c} models deterministically fails because many of the policy functions' estimated probabilities for issuing an alert are below 0.5, so always taking the action with the larger probability dramatically reduces the number of alerts issued---in some counties to zero. Note that policies which do not use all the alerts in their budget are penalized by the specification of our rewards model, resulting in \textsc{det.trpo.qhi} and \textsc{det.a2c.qhi} %
no longer outperforming \textsc{nws} with statistical significance across the 30 counties. Inclusion of future information mitigates this issue, but the results are not nearly as good as their stochastic counterparts.

Our big picture intuition for this result is that we are in a heavily data-constrained setting, which frustrates RL's ability to completely identify alert fatigue. The stochastic models address this by sampling alert days, inducing less alert streaks. Thought a bit unsatisfying, we believe this is an acceptable compromise given that, as mentioned in the main text, an online heat alert-RL system would likely utilize stochastic exploration anyway, to update itself over time. 

\begin{table*}[ht]
\small %
\centering
\begin{tabular}{l p{1.9cm}p{1cm}p{1.3cm}p{4cm}p{3cm}} 
  \hline 
 
&&&&\multicolumn{2}{c}{NOHR Hosps Saved vs NWS (vs Zero) Per Summer}\\
\cmidrule{5-6}
Policy & Median Diff. & WMW & P-value & Median / 10,000 Medicare & Approximate Total \\ 
  \hline
    *\textsc{topk} & 0.022 & 406 & 0.0002 & 0.023 (1.96) & 113 (9,603) \\ 
   \hline
  \textsc{random} & -0.015 & 177 & 0.88 & -0.013 (1.98) & -64 (9,697) \\ 
  \textsc{basic.nws} & -0.286 & 30 & 1.0 & -0.283 (1.45) & -1,388 (7,110) \\ 
  \textsc{aa.qhi} & 0.03 & 348 & 0.0025 & 0.029 (1.99) & 143 (9,747) \\ 
  \textsc{dqn} & -0.123 & 43 & 0.99995 & -0.102 (1.83) & -502 (8,989) \\ 
  \textsc{qrdqn} & -0.117 & 51 & 0.99991 & -0.098 (1.85) & -480 (9,046) \\ 
  \textsc{trpo} & -0.065 & 97 & 0.99742 & -0.063 (1.9) & -309 (9,321) \\ 
  \textsc{a2c} & -0.063 & 100 & 0.99689 & -0.062 (1.9) & -301 (9,317) \\ 
  \textsc{dqn.qhi} & 0.03 & 370 & 0.00242 & 0.029 (1.99) & 142 (9,746) \\
  \textsc{qrdqn.qhi} & 0.035 & 344 & 0.01121 & 0.031 (1.99) & 153 (9,767) \\ 
  \textsc{trpo.qhi} & 0.038 & 338 & 0.0154 & 0.038 (2.02) & 185 (9,897) \\ 
  \textsc{a2c.qhi} & \textbf{0.042} & 344 & 0.01121 & 0.045 (2.02) & 222 (9,902) \\ 
  \hline
  \textsc{dqn.f} & -0.417 & 40 & 0.99996 & -0.401 (0.49) & -1,966 (2,408) \\ 
  \textsc{qrdqn.f} & -0.42 & 48 & 0.99993 & -0.42 (0.47) & -2,059 (2,322) \\ 
  \textsc{trpo.f} & -0.062 & 103 & 0.99625 & -0.061 (1.9) & -297 (9,322) \\ 
  \textsc{a2c.f} & -0.063 & 101 & 0.99669 & -0.062 (1.9) & -302 (9,304) \\ 
  \textsc{dqn.qhi.f} & -1.895 & 1 & 1 & -1.901 (0) & -9,317 (0) \\ 
  \textsc{qrdqn.qhi.f} & 0.014 & 238 & 0.45904 & 0.012 (1.61) & 61 (7,876) \\ 
  \textsc{trpo.qhi.f} &  \textbf{0.046} & 345 & 0.01062 & 0.048 (2.02) & 236 (9,886) \\ 
  \textsc{a2c.qhi.f} & 0.036 & 343 & 0.01183 & 0.038 (2.02) & 187 (9,915) \\ 
  \hline
  \textsc{det.trpo.qhi} & -0.662 & 57 & 0.99985 & -0.579 (0) & -2,835 (2) \\ 
   \textsc{det.a2c.qhi} & -0.886 & 90 & 0.99837 & -0.794 (0.16) & -3,888 (807) \\ 
   \textsc{det.trpo.qhi.f} &  \textbf{0.032} & 271 & 0.21723 & 0.034 (1.83) & 168 (8,948) \\ 
   \textsc{det.a2c.qhi.f} & 0.02 & 287 & 0.13335 & 0.021 (2.01) & 104 (9,836) \\ 
   \hline
\end{tabular}
\caption{Extension of Table \ref{tab:main_results} to illustrate the comparisons in terms of absolute hospitalizations. Comparison between the average return of each counterfactual policy and that of the \textsc{nws} policy on the evaluation years, summarized across counties (e.g. ``Median Diff." is the median difference in average return). The last two columns also provide a comparison to the counterfactual with zero heat alerts issued. WMW is the Wilcoxon-Mann-Whitney statistic (higher is better); its associated p-value is also included.
The first policy, marked by *, requires oracle knowledge of the future weather. The approximate total number of hospitalizations is based on 49 million Medicare enrollees in 2011 \citeA{cms_stats_2011}.}   \label{tab:extended_table1} 
\end{table*}

\mysubsection{Additional Figures from the Post-hoc Contrastive Analysis}

\begin{figure*}[ht] 
\centering
\includegraphics[width=0.85\textwidth]{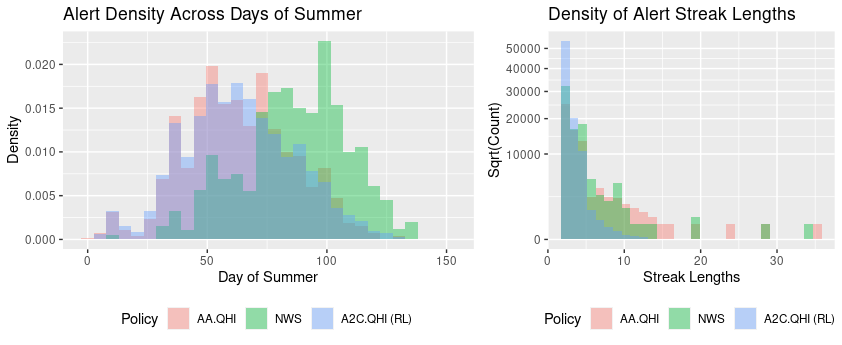}
\caption{Distributions of heat alert characteristics under different policies, over the evaluation years.}
\label{fig:hist_dos}
\end{figure*}

\begin{figure*}[ht] 
\begin{center}
\includegraphics[width=0.7\textwidth]{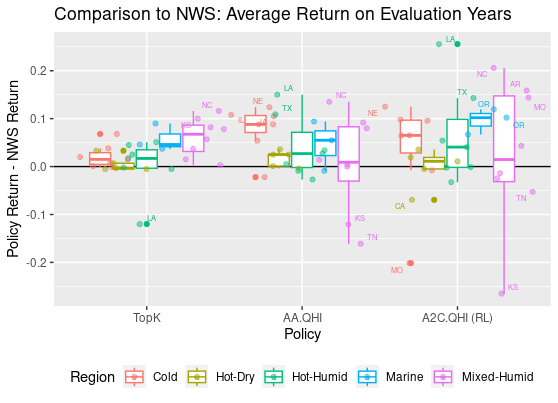}
\end{center}
\caption{Different policies' average return relative to \textsc{nws} across different climate regions, over the evaluation years. Outlier points are labeled with their state abbreviation. Note that \textsc{topk} is an oracle policy.}\label{fig:boxplot}
\end{figure*}

\begin{figure*}[ht] 
\begin{center}
\includegraphics[width=0.8\textwidth]{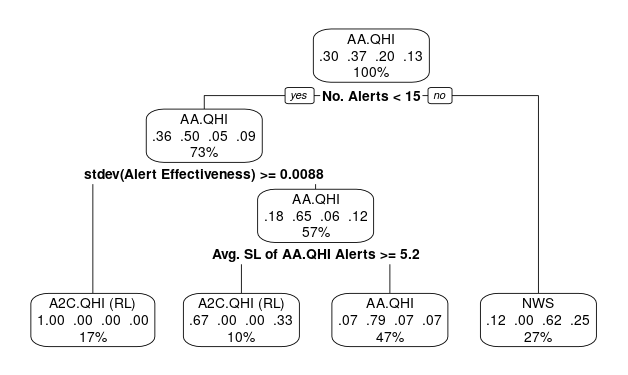}
\end{center}
\caption{Polytomous classification tree comparing \textsc{a2c.qhi}, \textsc{aa.qhi}, \textsc{nws}, and \textsc{trpo.qhi} across the 30 counties. Within each node, the policy at the top is that associated with the highest probability of performing best. The four probabilities beneath correspond to the four policies in alphabetical order, as written above. (Note that the two RL policies tend to be more similar to one another, as do the \textsc{aa.qhi} and \textsc{nws} policies, so the distribution of probabilities is important to consider in addition to the overall classification of each node.) The percentage at the bottom is the fraction of counties represented by that node. ``No. Alerts" is the average number of heat alerts per warm season during the evaluation years, ``Alert Effectiveness" is the $\tau$ estimated by our rewards model, and SL is the streak length.}
\label{fig:cart_nws_main}
\end{figure*}

\begin{figure*}[ht] 
\begin{center}
\includegraphics[width=0.8\textwidth]{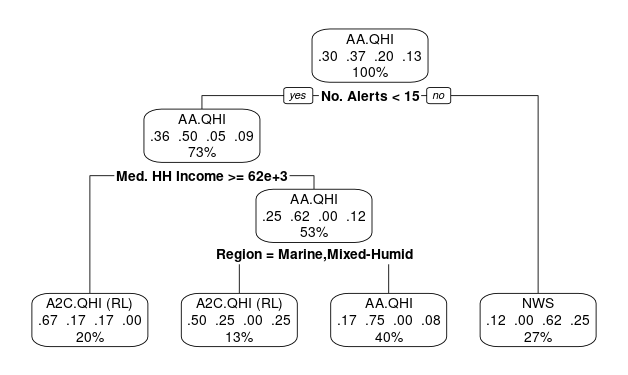}
\end{center}
\caption{Polytomous classification tree comparing \textsc{a2c.qhi}, \textsc{aa.qhi}, \textsc{nws}, and \textsc{trpo.qhi} across the 30 counties---using more conventional features than Figure \ref{fig:cart_nws_main}. Within each node, the policy at the top is that associated with the highest probability of performing best. The four probabilities beneath correspond to the four policies in alphabetical order, as written above. (Note that the two RL policies tend to be more similar to one another, as do the \textsc{aa.qhi} and \textsc{nws} policies, so the distribution of probabilities is important to consider in addition to the overall classification of each node.) The percentage at the bottom is the fraction of counties represented by that node. ``No. Alerts" is the average number of heat alerts per warm season during the evaluation years.} \label{fig:cart_interpretable}
\end{figure*}

Figure \ref{fig:hist_dos} provides histograms of different policies' temporal characteristics, figure \ref{fig:boxplot} compares different policies' performance (compared to NWS) across climate regions and counties, and figures \ref{fig:cart_nws_main}, \ref{fig:cart_interpretable}, and \ref{fig:cart_reg} illustrate the CART analysis.

\clearpage
\bibliographystyleA{aaai25}
\bibliographyA{all_refs}

\end{document}